\documentclass{article}
\usepackage[utf8]{inputenc}
\usepackage{graphicx}
\usepackage{amsmath}
\usepackage{amssymb}
\usepackage{booktabs}
\usepackage{float}
\usepackage{algorithm}
\usepackage{algpseudocode}
\DeclareMathOperator*{\argmax}{arg\,max}
\usepackage{lipsum}
\usepackage{multirow}
\usepackage{forest}
\usepackage[pagebackref,breaklinks,colorlinks]{hyperref}

\usepackage[capitalize]{cleveref}
\crefname{section}{Sec.}{Secs.}
\Crefname{section}{Section}{Sections}
\Crefname{table}{Table}{Tables}
\crefname{table}{Tab.}{Tabs.}
\graphicspath{{./}}
\usepackage{subfig}
\title{White-Box Attacks on Hate-speech BERT Classifiers in German with Explicit and Implicit Character Level Defense}

\author{Shahrukh Khan  \\
\texttt{shkh00001@uni-saarland.de}\\
\and
Mahnoor Shahid\\
\texttt{mash00001@uni-saarland.de}\\
\and
Navdeeppal Singh \\
\texttt{s8nlsing@stud.uni-saarland.de} \\
\url{https://github.com/shahrukhx01/adversarial-bert-german-attacks-defense}
}
\date{February 2022}

\begin{document}

\maketitle

\begin{abstract}
     In this work, we evaluate the adversarial robustness of BERT\cite{bert}
models trained on German Hate Speech datasets. We also complement our evaluation with
two novel white-box character and word level attacks thereby contributing to the range of attacks
available. Furthermore, we also perform a comparison of two novel character level defense
strategies and evaluate their robustness with one another.
\end{abstract}

\section{Introduction}
\label{sec:intro}

Natural language processing has achieved tremendous progress in surpassing
human-level baselines in a plethora of language tasks with the help of Attention based neural
architectures \cite{vaswani2017attention}.
However, recent studies\cite{hsieh-etal-2019-robustness, DBLP:journals/corr/abs-2004-01970, DBLP:journals/corr/abs-1905-11268} show that such neural models trained via transfer learning are susceptible to adversarial
noise. 
Moreover, to the best of our knowledge, most work concentrates on English language corpora. Hence,
in this study, we evaluate our attack and defense methods on BERT based models trained on German language hate-speech datasets.

\section{Background}
Hsieh et al.\cite{hsieh-etal-2019-robustness} proposed using self attention
scores for computing token importances in order to rank potential candidate tokens for perturbation.
However, one potential shortcoming of their idea is they replace the potential token candidate with random tokens from vocabulary which may result in changing the semantic meaning of
perturbed sample. 
Garg et al.\cite{DBLP:journals/corr/abs-2004-01970} proposed  BERT-based Adversarial Examples for Text Classification in which they employ Mask Language Modelling for generating potential word replacements in a black-box setting. 
Finally, Pruthi et al.\cite{DBLP:journals/corr/abs-1905-11268} showed susceptibility of BERT\cite{bert} based models to character-level miss-spellings also in a black-box setting. 
In our study, we employ both character-level and word-level attacks in a white-box setting.

\section{Experimental Setting}

\subsection{Undefended Models}
\subsubsection{Datasets}
We present our work based on HASOC 2019 (German Language) \cite{10.1145/3368567.3368584} and GermEval 2021 \cite{risch-etal-2021-overview}  sub-task 1 respectively. Both of the sub-tasks are binary classification tasks where the positive labels correspond to hate-speech and negative labels correspond to non-hate-speech examples.
\begin{table}[H]
  \centering
  \begin{tabular}{@{}cc@{}cc@{}cc@{}}
    \toprule
    Dataset & Train & Validation & Test \\
    \midrule
    HASOC 2019 & 3054 & 765 & 850  \\
    GermEval 2021 & 2594 & 650 & 944\\
    \bottomrule
  \end{tabular}
  \caption{Dataset Statistics }
  \label{tab:example1}
\end{table}
\subsubsection{Training}\label{sec:std_training}
For training, the undefended models, we fine-tune the GBERT\cite{DBLP:journals/corr/abs-2010-10906} language model for German language which employs training strategies namely \emph{Whole Word Masking} (WWM) and evaluation driven training and currently achieves SoTA performance for document classification task for German language. We obtain the following accuracy scores for each dataset respectively.

\begin{table}[H]
  \centering
  \begin{tabular}{@{}cc@{}cc@{}cc@{}}
    \toprule
    Dataset & Accuracy(\%) \\
    \midrule
    HASOC 2019 & 84 \\
    GermEval 2021 & 69 \\ 
    \bottomrule
  \end{tabular}
  \caption{Undefended Models }
  \label{tab:example2}
\end{table}

\subsection{Attacks}

\subsubsection{Baseline Word-level White-Box Attack}

The baseline word-level attack is composed by enhancing Hsieh et. al\cite{hsieh-etal-2019-robustness} which prominently replaces tokens sorted in order of their attention scores with random tokens from vocabulary which may lead to perturbed sequence being semantically dissimilar to the source sequence. 
In the baseline attack, we address this potential shortcoming by using a language model using \emph{Masked Language Modeling} (MLM) to generate potential candidate for each token ranked in the order of attention scores. 
Furthermore, instead of just performing the replacement operation, we employ the perturbation scheme as proposed by Garg et al.\cite{DBLP:journals/corr/abs-2004-01970} we insert generated tokens to left/right of the target token where the candidate tokens are generated via MLM.

\subsubsection{Word-level White-Box Attack }

The main motivation behind this attack is based on the fact using only language models to ensure semantic correctness in the adversarial sequences is not enough. 
Since it highly depends on the vocabulary of the pre-trained language model. 
We improve the baseline attack for the preserving more semantic and syntactic correctness of the source sequence by introducing further constraints on the generated sequence by the baseline attack. 
Firstly, we compute the document-level embeddings for both perturbed and source sequence and then compute cosine similarity with a minimum acceptance threshold of \textbf{0.9363} as originally suggested by Jin et al.\cite{DBLP:journals/corr/abs-1907-11932}, since Garg et al.\cite{DBLP:journals/corr/abs-2004-01970} developed their work using the same threshold value.
Finally, we further add another constraint that Part of Speech (POS) tag of both candidate and target token should be same. 

\subsubsection{Character-level White-Box attack }

In this white-box character level attack, similar to earlier white-box word-level attacks attention scores are obtained in order to get the word importance. Then, by ordering the word importance in the order of higher to lower we employ the character perturbation scheme employed by Pruthi et al.\cite{DBLP:journals/corr/abs-1905-11268} since they evaluated this in the black-box setting only, we perform character level perturbation within a target token by token modification of character (swap, insert, delete etc) applied to cause perturbations such adversarial examples are utilized to maximize the change in model’s original prediction confidence with limited numbers of modifications. However, these modifications prove to be significantly effective as outlined in the results section.

\subsection{Defenses}
\subsubsection{Abstain Based Training}
In several past evaluations and benchmarks of defenses against adversarial examples \cite{Athalye2018ObfuscatedGG,Carlini2017AdversarialEA,Carlini2017TowardsET,DBLP:conf/icml/Croce020a,DBLP:journals/corr/croceRobustBench2020,DBLP:journals/corr/BryniarskiEvadingDetectionMethod2021}, adversarial training \cite{Madry2018TowardsDL} has been found to be one of the best ways of conferring robustness.
However, it is computationally expensive due to the need of creating adversarial examples during training. 
Thus, we chose to employ a detection based defense, which we call \textit{abstain based training}. 
Although, detection based defenses are known to be not as effective as adversarial training \cite{Carlini2017AdversarialEA,DBLP:journals/corr/BryniarskiEvadingDetectionMethod2021}, we still believe our method will deliver insights into the capability of BERT models in recognizing adversarial examples similar to adversarial training due the way it works.
In contrast to other detection based defenses in the literature \cite{DBLP:journals/corr/GrosseMP0M17,DBLP:journals/corr/GongWK17,DBLP:conf/cvpr/BendaleB16,Sotgiu2020DeepNR,DBLP:conf/iclr/MetzenGFB17}, the approach is much simpler.
It works as follows. 

\begin{figure}[H]
    \centering
        \begin{tabular}{c}
            \includegraphics[width=\linewidth]{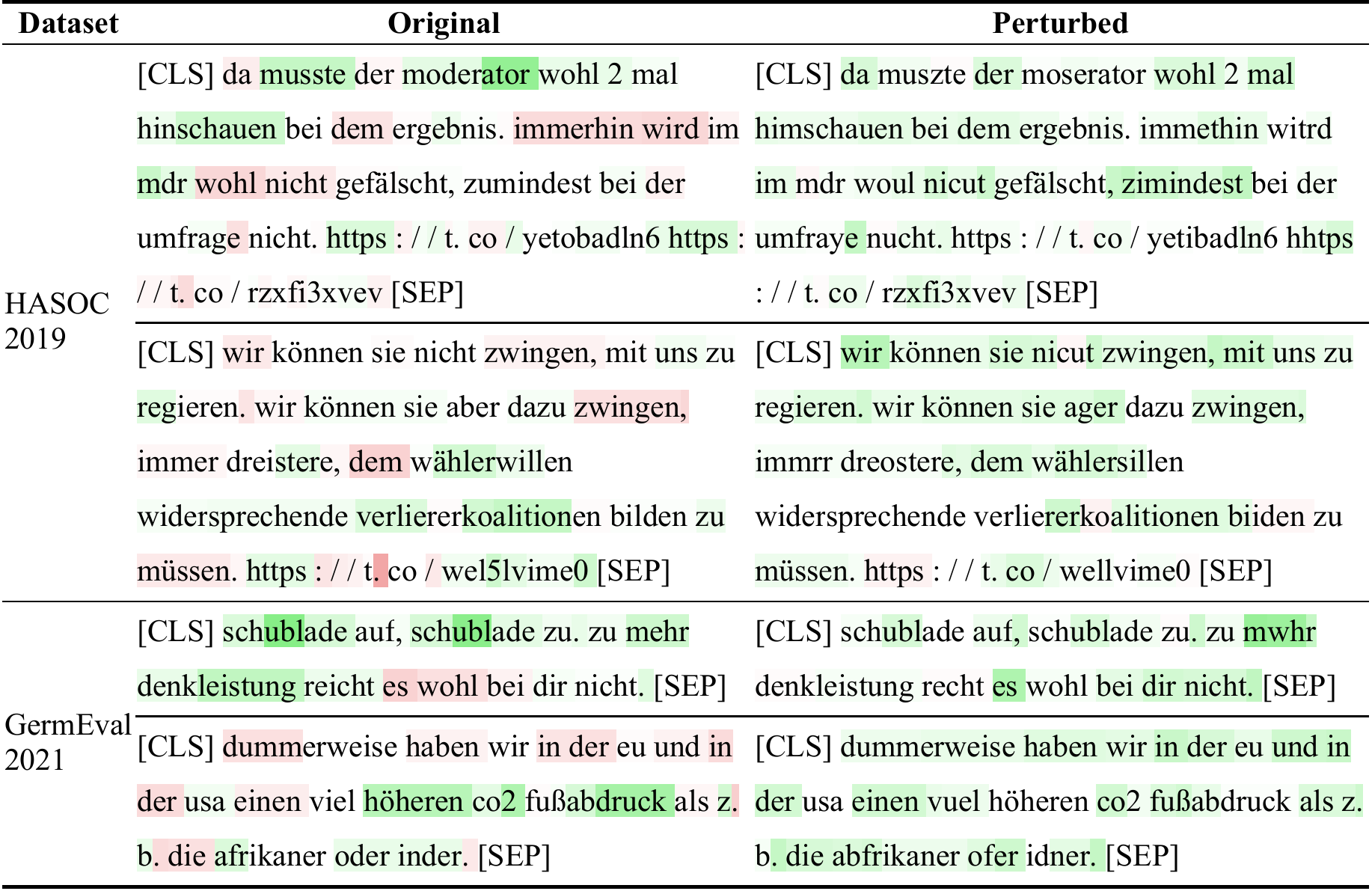}
        \end{tabular}
    \caption{Visualization of the classification attributions of the abstain based trained models, which correctly classify the examples. The \emph{perturbed} examples shown above fool the normally trained models. We observe that the attributions are much more spread out when models encounters a perturbed example. (Words were split by the tokenizer, thus a single word can have different sub-attributions.)}
    \label{tab:at_scores_abstain}
\end{figure}

Let $C$ be the trained undefended classifier.
We create a new (untrained) classifier $C'$ from $C$ by extending the number of classes it is able to predict by one. 
The new class is labeled `ABSTAIN', representing that the classifier abstains from making a prediction.
Using $C$ we create the adversarial examples. 
We mix these with the normal examples from the dataset (of $C$), where the adversarial examples have the abstain label, to create a new dataset.
We then simply train on this dataset.
We applied this defense strategy on the models from \cref{sec:std_training} and present the results in \cref{tab:example}. We also show the classification attributions in \cref{tab:at_scores_abstain} to try to interpret the models' behaviour.

\subsubsection{Explicit Character-Level Defense}
Abstain based training defense achieves high success in defending against the adversarial character-level perturbed inputs. 
However, this results in degraded system utility since the model does not make any useful prediction when the input is perturbed at character-level. 
To overcome this drawback, we propose the explicit character-level defense which is an unsupervised approach which makes an assumption that 
\[
 {\forall} t \in T_{input}: \quad t \in V_{train}\ .
\]
Here, $V_{train}$ is the set of all tokens present in the training set. However, replacing this set with set of words in the given language i.e., set of all words in German language etc. would result in better results. $T_{input}$ refers to set of tokens present in the input sequence and we assume the worst case which means $T_{input}$ is perturbed with character-level noise.

\begin{figure}[H]
\centering
  \includegraphics[scale=0.7]{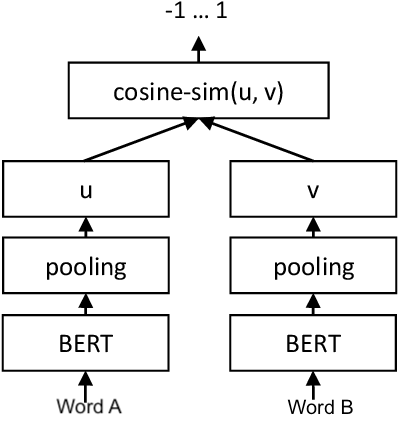}
  \caption{Sentence-BERT for character-level similarity.}
  \label{fig:sbert}
\end{figure}

In this defense method, we firstly re-purpose the Sentence-BERT\cite{reimers-2019-sentence-bert} architecture which originally trained sentence pairs to compute semantic vector representations and achieved SoTA results on multiple Information retrieval datasets. However, we change input to character level by inputting word pairs to the network. 
Concretely, we labelled the Birkbeck spelling error corpus\cite{20.500.12024/0643} which has word pairs with one correct and the other misspelled word and we label the each pair based on the Levenshtein distance between each pair. 
The schematics of our neural approach are given in \cref{fig:sbert}.

The main idea behind using the neural approach is to project similarly spelled words close to each other in the vector space. \cref{alg:explicit_defense} outlines main idea of our approach for explicit character-level defense.

\begin{algorithm}[H]
\caption{Explicit Character Level Defense}\label{alg:cap}
\begin{algorithmic}
\State $V_{train} \gets t_{1} \dots t_{m}$ \Comment{Set of tokens in vocabulary}
\State $E_v \gets \Vec{e_{1}} \dots \Vec{e_{m}}$ \Comment{Embeddings of vocabulary} \newline
$T_{input} \gets t_{1} \dots t_{j} $ \Comment{Set of tokens in input}

\For{$k \gets 1$ to $j$}
        \State $\Vec{e_{k}} \gets v_{1} \dots v_{n} $ \Comment{Get embedding of input token k} 
        \State$\Vec{scores} \gets  cos(E_v,  \Vec{ e_{k}}) $
        
        \Comment{Cosine similarity with vocabulary embeddings}

         {
        \If{$\max \Vec{scores} \geq  0.7 \quad and \quad\max \Vec{scores} < 1.0 \quad$}
        {
            $vocab_{index} \gets \argmax{\Vec{scores}}$;
            
            $\quad T_{input}[k] \gets V_{train}[vocab_{index}]$
        }
    \newline
}
      \EndFor
\end{algorithmic}
\label{alg:explicit_defense}
\end{algorithm}

\section{Results}
\subsection{Attack Results}

As shown in \cref{tab:example_attack} character-level attacks prove to be most effective on both models.
\begin{table}[H]
  \centering
  \begin{tabular}{@{}cc@{}cc@{}cc@{}cc@{}}
    \toprule
    Dataset & Attack  & Success rate(\%) \\
    \midrule
    HASOC 2019 & \multirow{2}{*}{Baseline} &  8.49  \\
    GermEval 2021 &  & 60.3\\
    HASOC 2019 & \multirow{2}{*}{Word-level}   & 4.03\\
    GermEval 2021 & &  49.8\\
    HASOC 2019 & \multirow{2}{*}{Character-level} & \textbf{73.1}\\
    GermEval 2021 && \textbf{93.5}\\

    \bottomrule
  \end{tabular}
  \caption{Attacks Result on Undefended Models.}
  \label{tab:example_attack}
\end{table}

\begin{figure}[H]
\vspace*{-1.5in}
\hspace*{-0.5in}
\begin{tabular}{cc}
\centering
\subfloat[\label{fig:avg_queries}]{\includegraphics[width = 3.0in]{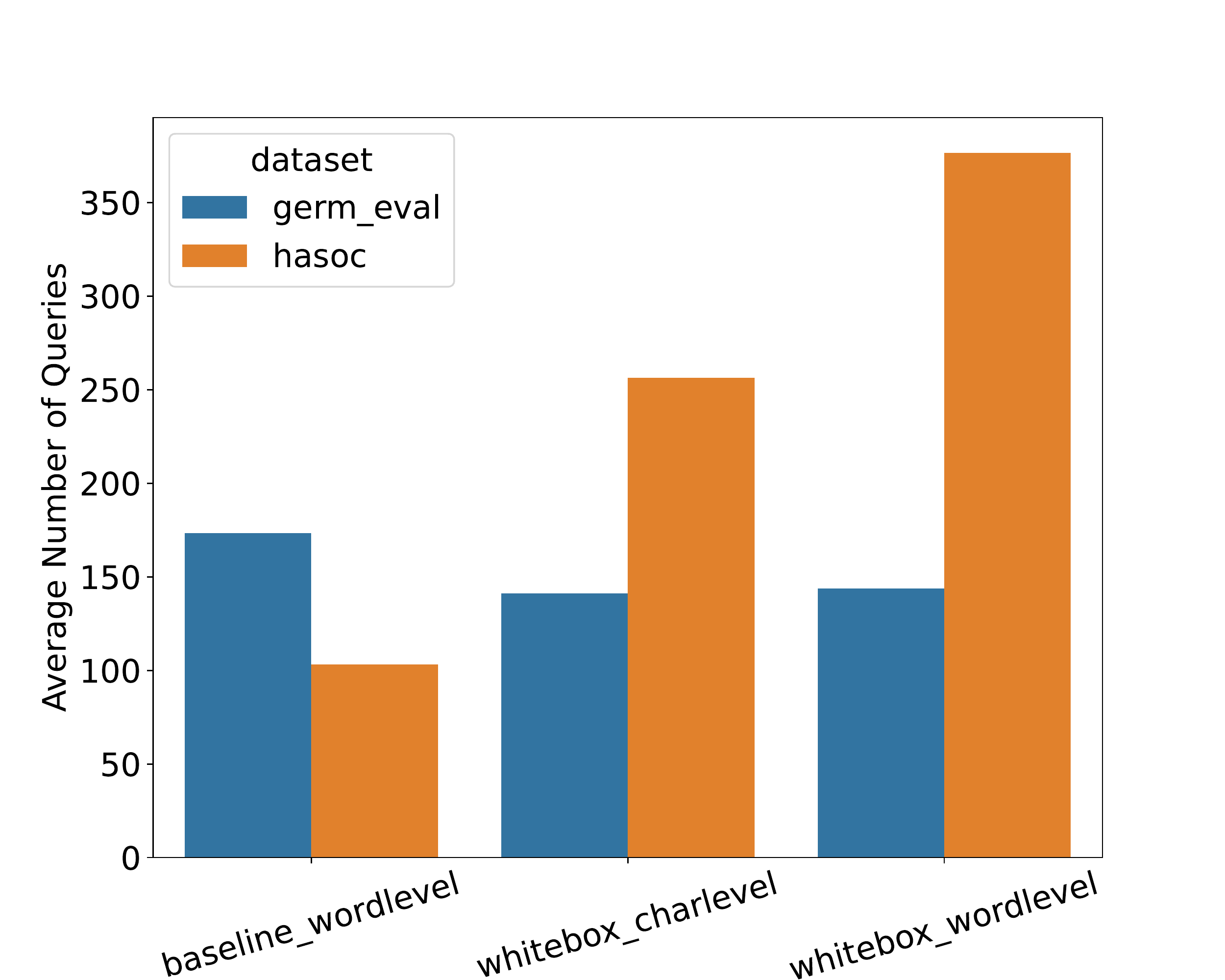}} &
\subfloat[\label{fig:pearson_len_query}]{\includegraphics[width = 3.0in]{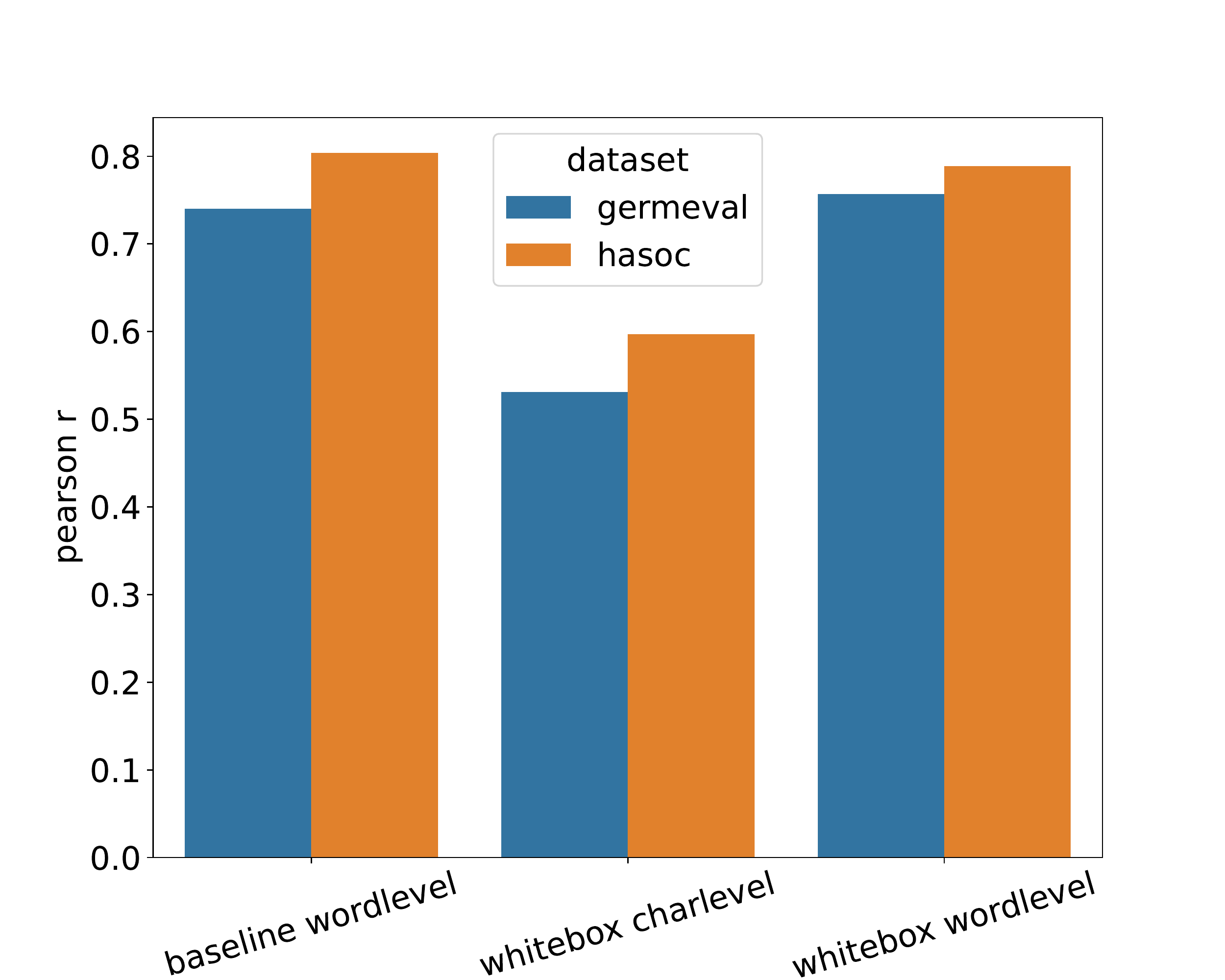}} \\
\subfloat[\label{fig:levenshtein_dist}]{\includegraphics[width = 3.0in]{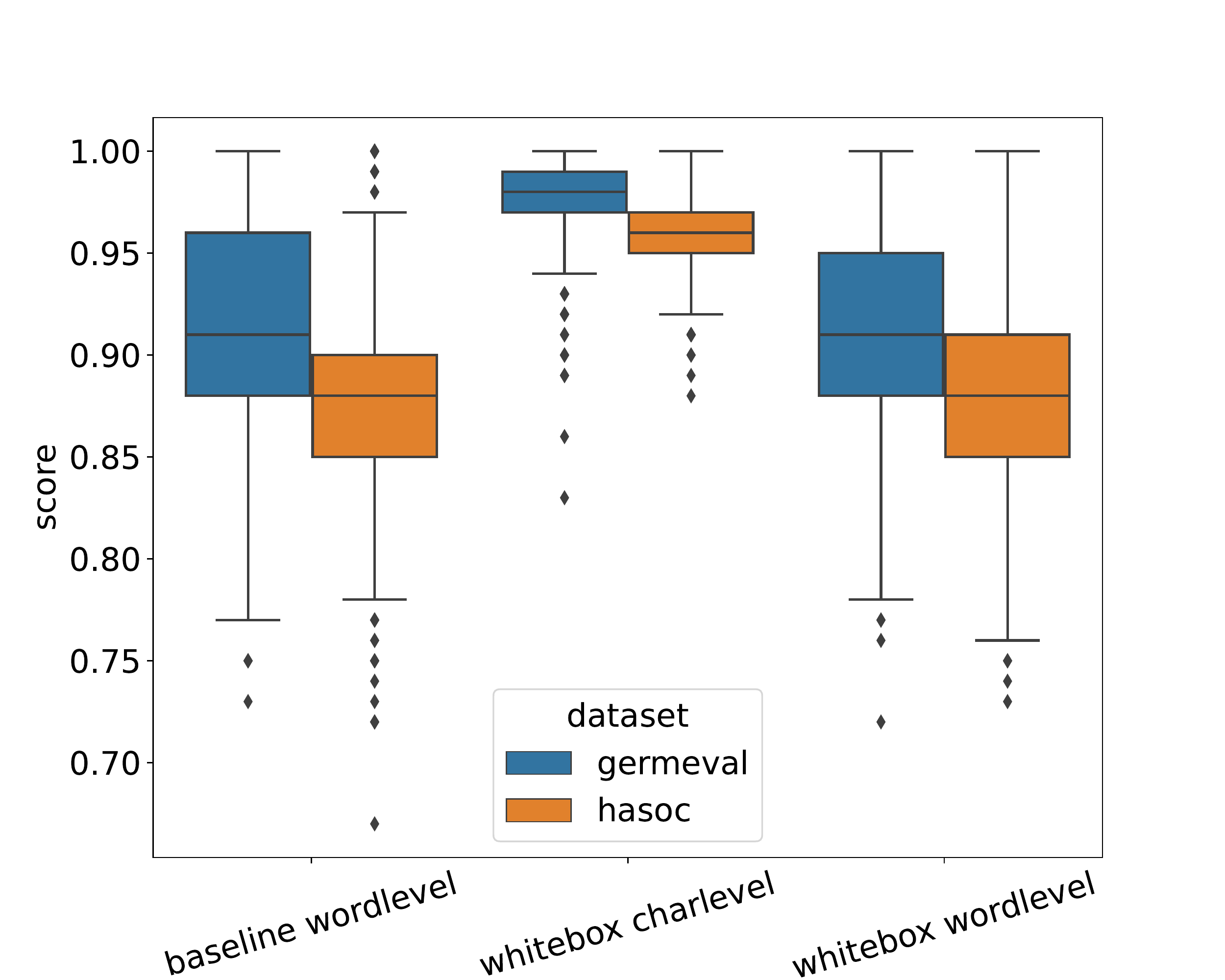}} &
\subfloat[\label{fig:confidence_delta}]{\includegraphics[width = 3.0in]{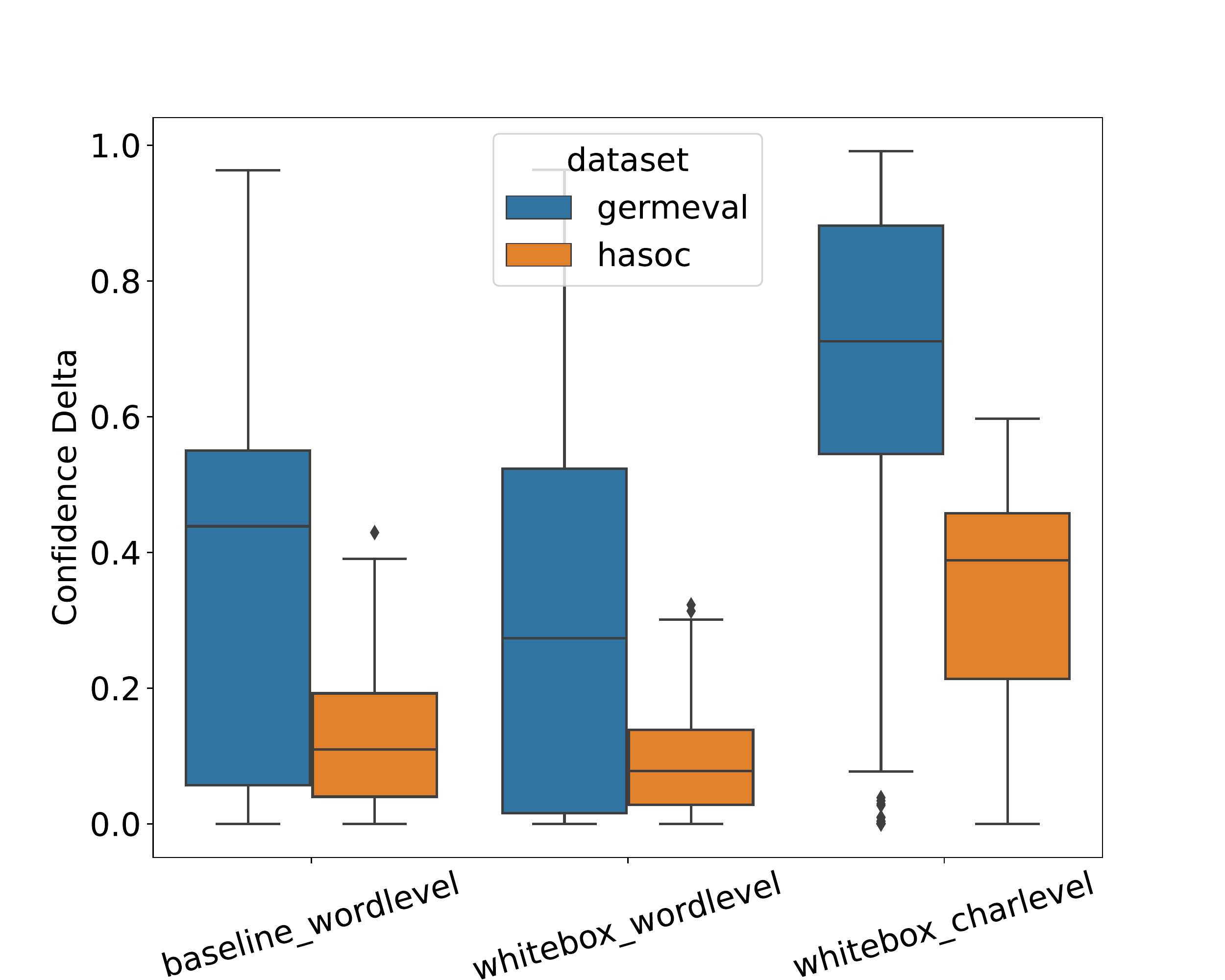}}
\end{tabular}
\caption{\newline \cref{fig:avg_queries} illustrates how number of queries required per sample for a successful attack depends on the dataset and attack type, we further show in  \cref{fig:pearson_len_query} that both word-level attacks require more queries for a longer sequence as compared to character-level attack which is slightly agnostic to the sequence length. \cref{fig:levenshtein_dist} shows that the character level attack require minimal amount of perturbation since the changes are at word level, moreover from \cref{fig:confidence_delta} it can be concluded that character-level attack also makes the highest difference in model prediction confidence in case of a successful attack.}
\end{figure}

\subsection{Defense Results}
\begin{table}[H]
  \centering

  \begin{tabular}{@{}cc@{}cc@{}cc@{}cc@{}}
    \toprule
    Dataset & Defense  & Attack success rate(\%) \\
    \midrule
    HASOC 2019 & \multirow{2}{*}{Explicit Character Level} &  9.5  \\
    GermEval 2021 &  & \textbf{5.3}\\
    HASOC 2019 & \multirow{2}{*}{ Implicit Abstain-based }   & \textbf{1}\\
    GermEval 2021 & &  11.1\\

    \bottomrule
  \end{tabular}
  
  \caption{Character-level Attack on Defended Models  }
  \label{tab:example}
\end{table}

\section{Conclusion}
\begin{figure}[H]
\centering
  \includegraphics[scale=0.35]{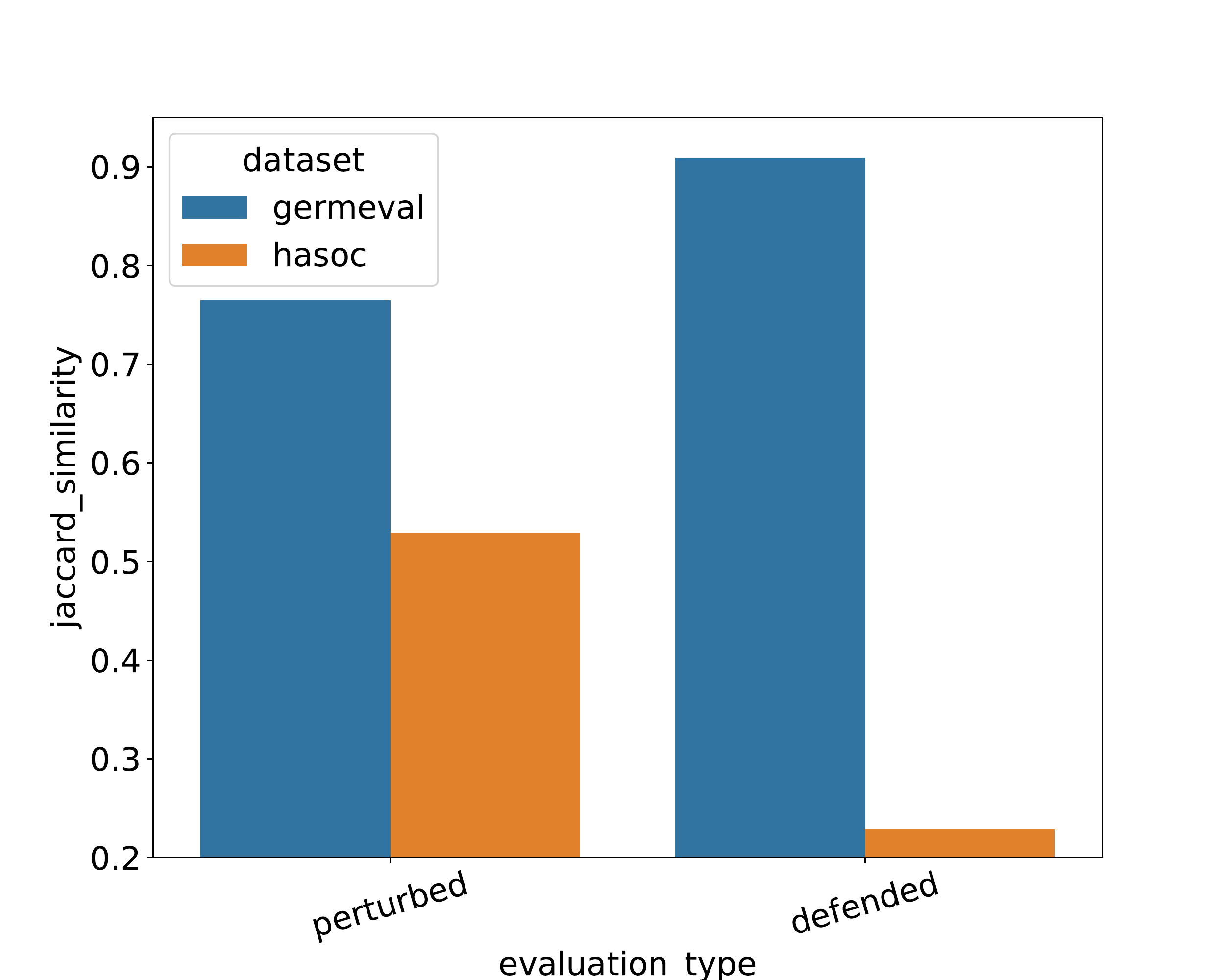}
  \caption{Jaccard similarity between original and perturbed text vs.~the original and defended text.}
  \label{fig:jaccard_sim}
\end{figure}

We show that self-attentive models are more susceptible to character-level adversarial attacks than word-level attacks on  text classification NLP task.
We provide two potential ways to defend against character-level attacks. 
Future work can be done to enhance the explicit character-level defense using supervised sequence-to-sequence neural approaches, since as shown in \cref{fig:jaccard_sim} current approach enhance the jaccard similarity of defended sequences with original sequences when compared to jaccard similarity between original sequence and perturbed sequence in case of GermEval 2021.
However, for HASOC 2019 dataset because of abundance of Out of Vocabulary tokens in the unseen test set the defense degrades the quality of defended sequences.
However, even then the defense proves to be quiet robust against character-level adversarial examples as shown in \cref{tab:example}
{\small
\bibliographystyle{unsrt}
\bibliography{egbib}
}

\end{document}